\documentclass[10pt,twocolumn,letterpaper]{article}
\usepackage{cvpr}            
\usepackage{graphicx}
\usepackage{amsmath}
\usepackage{amssymb}
\usepackage{booktabs}
\usepackage{float}
\usepackage{url}
\usepackage{adjustbox}
\usepackage{bm}
\usepackage{array}
\usepackage{multirow}
\usepackage{subcaption}
\usepackage{pifont}
\usepackage{placeins}
\usepackage{makecell}
\usepackage{newfloat}
\usepackage{listings}
\usepackage{bibentry}
\usepackage{algorithmic}
\usepackage{xcolor}
\usepackage[accsupp]{axessibility}
\makeatletter
\newif\if@restonecol
\makeatother

\usepackage[ruled,vlined]{algorithm2e}

\usepackage[pagebackref,breaklinks,colorlinks]{hyperref}

\usepackage[capitalize]{cleveref}
\crefname{section}{Sec.}{Secs.}
\Crefname{section}{Section}{Sections}
\Crefname{table}{Table}{Tables}
\crefname{table}{Tab.}{Tabs.}

\def\cvprPaperID{362} 
\def\confName{WACV}
\def\confYear{2022}

\begin{document}

\title{SD-Conv: Towards the Parameter-Efficiency of Dynamic Convolution}

\author{    
Shwai He\textsuperscript{\rm 1, \rm 2} \and
Chenbo Jiang\textsuperscript{\rm 3}\and
Daize Dong\textsuperscript{\rm 2}\and
Liang Ding\textsuperscript{\rm 1}\thanks{Corresponding author}\and
    \textsuperscript{\rm 1}JD Explore Academy\\
    \textsuperscript{\rm 2}University of Electronic Science and Technology of China\\
    \textsuperscript{\rm 3}Nanjing University of Science and Technology Zijin College\\
    {\tt\small shwai.he@gmail.com},
    {\tt\small cbjiang@foxmail.com}, \\
    {\tt\small dzdong2019@gmail.com},
    {\tt\small dingliang1@jd.com}
}

\maketitle

\begin{abstract}
Dynamic convolution achieves better performance for efficient CNNs at the cost of negligible FLOPs increase. However, the performance increase can not match the significantly expanded number of parameters, which is the main bottleneck in real-world applications. Contrastively, mask-based unstructured pruning obtains a lightweight network by removing redundancy in the heavy network. In this paper, we propose a new framework, \textbf{Sparse Dynamic Convolution} (\textsc{SD-Conv}), to naturally integrate these two paths such that it can inherit the advantage of dynamic mechanism and sparsity. We first design a binary mask derived from a learnable threshold to prune static kernels, significantly reducing the parameters and computational cost but achieving higher performance in Imagenet-1K. We further transfer pretrained models into a variety of downstream tasks, showing consistently better results than baselines. We hope our SD-Conv could be an efficient alternative to conventional dynamic convolutions. 
\end{abstract}

\section{Introduction}
There have been rich discussions on the representation power of deep neural networks in two opposite directions~\cite{raghu2017expressive,lu2017expressive}. From the perspective of increasing the model capacity, more layers and channels with specialized infrastructure (e.g. dynamic convolution \cite{chen2020dynamic}) can achieve higher performance with less overfitting~\cite{nguyen2020wide, allenzhu2020learning}. In the view of model compression, network pruning and quantization of complex networks can induce smaller models possibly at the expense of minor accuracy loss~\cite{gale2019state, fan2021training}. Regarding the trade-off between cost and gain in these two opposite approaches, what will happen when we combine them for infrastructure optimization? Especially, can we combine the advances of dynamic convolution and sparsity towards the best of both worlds -- achieving a desirable trade-off between complexity and performance?

Dynamic convolution (DY-Conv)~\cite{chen2020dynamic} achieves significant performance gains over static convolution with negligible computational cost but relatively high memory cost. Specifically,  it utilizes an input-based attention mechanism to generate dynamic attention weights to combine multiple parallel static kernels, boosting the performance at the cost of increased convolutional parameters.

However, during inference, this parameter increase does not match the model performance improvement completely. For example, DY-ResNet-18 \cite{chen2020dynamic} is 2.3\% higher than ResNet-18 and 4\% lower than ResNet-50 in Top-1 accuracy on Imagenet-1K \cite{5206848}, while its parameter amount is about four times of ResNet-18 and twice of ResNet-50. In addition, large-scale DNNs require huge storage and deployment cost, which becomes the main bottleneck of the real-world deployment \cite{kusupati2020soft, frankle2019lottery, iofinova2022well}. These phenomena raise the problem of parameter efficiency when we try to adopt dynamic convolution more efficiently. 

A possible routine to improve the parameter efficiency in dynamic convolution is to refine the method of kernel combination. For example, Li~\textit{et al.} \cite{li2021revisiting} reformulate the linear combination of dynamic convolution into a summation of the static kernel and sparse dynamic residual. Another scheme is to increase the sparsity to build compact parameter-efficient networks. One can prune task-unrelated neurons that usually have small absolute values~\cite{han2015learning,zhu2017prune,frankle2019lottery} or contribute little to the decrease of loss function~\cite{pmlr-v97-peng19c, li2017pruning,8712713}. 

In recent studies, some sparse networks not only decrease storage and computational requirements but also achieve higher inference scores than dense networks \cite{rigl}, suggesting the potential utility of sparse structure in alleviating overfitting problems \cite{huang-etal-2022-sparse,XU201969}. In terms of the representation power of subnetworks, the Lottery Ticket Hypothesis \cite{frankle2019lottery} shows that there consistently exists lightweight subnetworks that can be trained from scratch with competitive learning speed as their larger counterparts, while maintaining comparable test accuracy. 
Based on this hypothesis, we assume that we can find the subnetworks for dynamic convolution in the training process and achieve a compact and efficient dynamic convolution network.

In this work, we re-examine the parameter efficiency for dynamic convolution. We first simply prune out half of the parameters in the $k$ parallel kernels for pretrained dynamic convolution layers. Surprisingly, we find that this pruning operation has minimal effect on the numerical feature of dynamic parameters and a negligible impact on performance.

Based on this discovery, we further propose to integrate dynamic convolution with sparsity, namely sparse dynamic convolution, which enjoys natural complementarity. Technically, we present a new algorithm to train the dynamic convolution modes via iterative pruning.  Specifically, we set a learnable threshold for each convolutional layer and prune the neurons whose magnitudes are below the threshold. We also propose a penalty term to explicitly regulate the $L_0$-norm of maintained parameters to guarantee the total parameters under an overall budget without additional hyperparameters. Considering that the computational kernel is a linear combination of static kernels, the masked kernels can then be integrated into a sparsely computing kernel with reduced FLOPs. 
\begin{figure*}[t]
\centering
\caption{Illustration of convolution kernel generation process for Dynamic Convolution (Left) and our proposed Sparse Dynamic Convolution architecture (Right). }
\begin{subfigure}[t]{0.495\textwidth}
\centering
\includegraphics[scale=0.16]{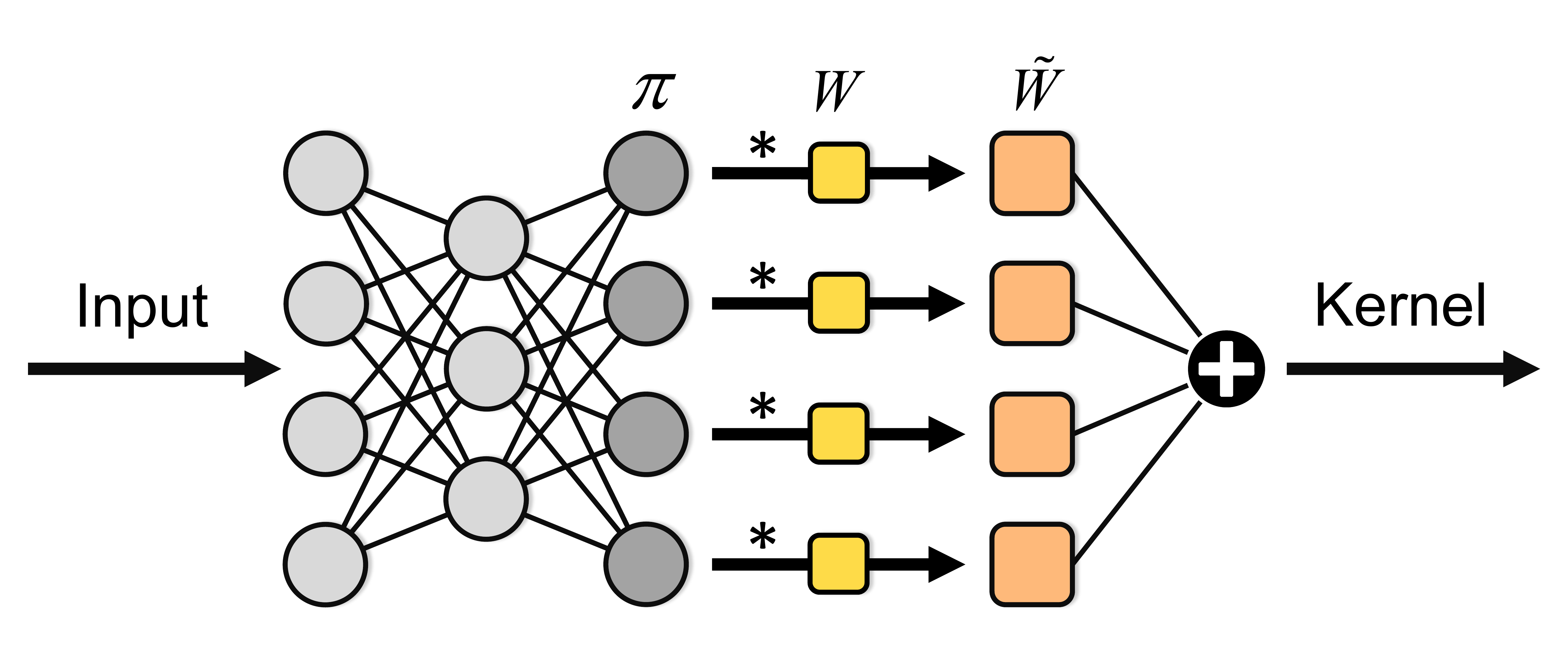}
\subcaption{Dynamic Convolution.}
\end{subfigure}
\hfill
\begin{subfigure}[t]{0.495\textwidth}
\centering
\includegraphics[scale=0.16]{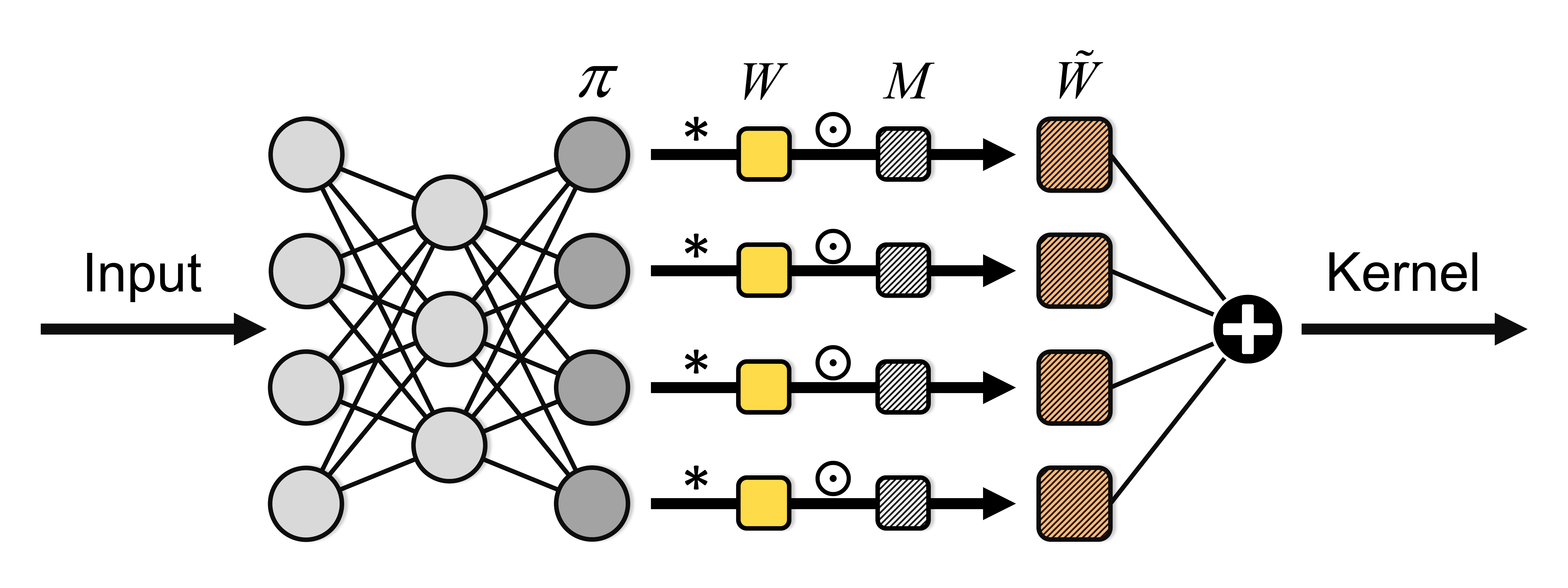}
\subcaption{Sparse Dynamic Convolution.}
\end{subfigure}
\vspace{-5pt}
\end{figure*}

To validate the efficiency of sparse dynamic convolution, we execute our methods on both ImageNet-1K \cite{5206848} and downstream tasks, and demonstrate the promotion that arises from both dynamic convolution and sparsity: Dynamic mechanism improves the representation power with negligible extra computational cost; Sparsity reduces the redundancy of dynamic convolution and promotes the performance during inference. In short, our main contributions are as follows: 
\begin{itemize}
\item[$\bullet$]We propose the Sparse Dynamic Convolution (SD-Conv) to improve the parameter efficiency of dynamic convolution by marrying the dynamic convolution and sparsity to maintain the advantage of both worlds.
\item[$\bullet$]We propose a novel $L_0$-norm based pruning method with an optimization policy to train sparse dynamic convolution networks efficiently. 
\item[$\bullet$]Our experiments on both upstream tasks and downstream tasks have shown the complementarity between sparsity and dynamic convolution. 
\end{itemize}

\section{Related Work}
Both dynamic convolution and sparsity are often considered separately to promote neural networks. We delve into the combination of them and briefly review them as follows:

\paragraph{Dynamic Networks} 
Dynamic networks adapt input-based parameters or activation functions to boost representation power. HyperNetworks\cite{ha2016hypernetworks} use a secondary network to generate parameters for the main network. SENet \cite{hu2019squeezeandexcitation} applies channel-wise attention to channels. DRConv\cite{chen2021dynamic} transfers the increasing channel-wise filters to a spatial dimension with a learnable instructor. 
CondConv\cite{yang2020condconv} and Dynamic Convolution \cite{chen2020dynamic} each proposed a new convolution operator to improve the representation capability with negligible extra FLOPs. Instead of using a single static convolution kernel per layer, they use the linear combination of a set of $k$ parallel static kernels $\{W_i, b_i\}\space( i = 1, 2, \dots, k$), where the linear scale is dynamically aggregated via a function of individual inputs. Dynamic convolution \cite{chen2020dynamic} utilizes an attention function to formulate the linear score: 
\begin{equation}
\begin{aligned}
    \hat W=&\sum_{i=1}^{k} \pi_{i} \cdot W_{i} \\
    \text{s.t.}\quad&\sum_{i=1}^{k}\pi_{i}=1,\quad 0\leq\pi_{i}\leq1,
\end{aligned}
\end{equation}

where $\pi_k$ is the attention score of the $k$-th kernel. Dynamic convolution only introduces two negligible additional computations:
1) Computing the attention scores $\pi_i\space( i = 1, 2, \dots k)$.
2) Aggregating parameters based on attention scores $\sum_{i=1}^k \pi_i(x) * W_i$. 
This linear combination significantly promotes the representation power of dynamic convolution and improves the performance in mainstream computer vision tasks. 

However, towards the use of $k$ parallel kernels in dynamic convolution, Li~\textit{et al.}~\cite{li2021revisiting} have proposed that it lacks compactness, and further utilized a matrix deposition method to improve this problem. Similarly, our work investigates the parameter efficiency of dynamic convolution and utilizes network pruning methods to improve it. 

\paragraph{Sparsity}
Sparsity has been widely studied to compress deep neural networks in resource-constrained environments. It can be generally categorized into two groups: \textsc{structured} and \textsc{unstructured} sparsity. Structured sparsity prunes blocks of sub-networks in a neural network, while unstructured fine-grained sparsity prunes multiple individual weights distributed across the whole neural network. Between the two sparsity types, unstructured sparsity usually achieves significantly higher compression ratios while maintaining relatively better performance \cite{han2015learning, guo2016dynamic}, which therefore leaves as our default sparsity type.

Unstructured sparsity usually detects unimportant parameters and utilizes a threshold to prune them. On the one hand, many previous works compute the threshold based on different importance-based criteria,  including magnitude \cite{han2015learning, zhu2017prune, frankle2019lottery}, Hessian-based heuristics \cite{li2017pruning, 8712713} and connection sensitivity \cite{lee2019snip, luo2020neural}. On the other hand, sparse training with differential thresholds has also been widely explored. Kusupati~\textit{et al.}~\cite{kusupati2020soft} and Manessi~\textit{et al.}~\cite{manessi2018automated} propose to learn layer-wise thresholds automatically using a soft thresholding operator or a close variant of it. As the learning-based thresholds contribute to the minimization of task-specific loss, the differential thresholds-based sparse method \cite{kusupati2020soft} contributes to high performance.
Besides, sparsity learned during training with approximate $L_0$-norm regulation has also been used in several works \cite{louizos2018learning, azarian2021learned}, because it controls the overall sparsity directly. To make the $L_0$-norm of thresholds differentiable, Louizos~\textit{et al.}~\cite{louizos2018learning} set a collection of non-negative stochastic gates to determine the weights to be pruned, while Azarian~\textit{et al.}~\cite{azarian2021learned} propose an approximate form of $L_0$-norm to estimate the gradient. Considering both performance and controllability, we adopt a threshold-based $L_0$-norm in our sparse method.

\section{Methodology}
In this section, we first present our motivation for sparsity in dynamic convolution, then illustrate our efficient architecture, namely Sparse Dynamic Convolution (\textsc{SD-Conv}). 

\subsection{Motivation}
In conventional dynamic convolution, each convolutional layer prepares $k$ parallel kernels to aggregate the dynamic kernel, leading to a nearly $k$ times larger model and potential parameter redundancy. For example, the total parameters of dynamic ResNet-50 (DY-ResNet-50) are about 100.9M (with 4 kernels) compared to about 23.5M for static ResNet-50. For this phenomenon, we raise two questions: 
(1) \textit{Is it necessary to pay the cost of enormous parameters and computations, e.g. 329\% in DY-ResNet-50, to aggregate the dynamic kernels?}
(2) \textit{Is it necessary to deploy all of these parameters to maintain the slight performance improvement, e.g. 1.1\% in DY-ResNet-50?} 

\begin{figure} [t!]
  \centering
    \caption{The scaled layer-wise mean and variance values of the aggregated kernel weights of ResNet-50. ``Vanilla'' and ``Pruned'' denote dynamic convolution networks before and after pruning, respectively. The mean and variance values keep nearly unchanged after pruning 50\% parameters. }
  \includegraphics[scale=0.55]{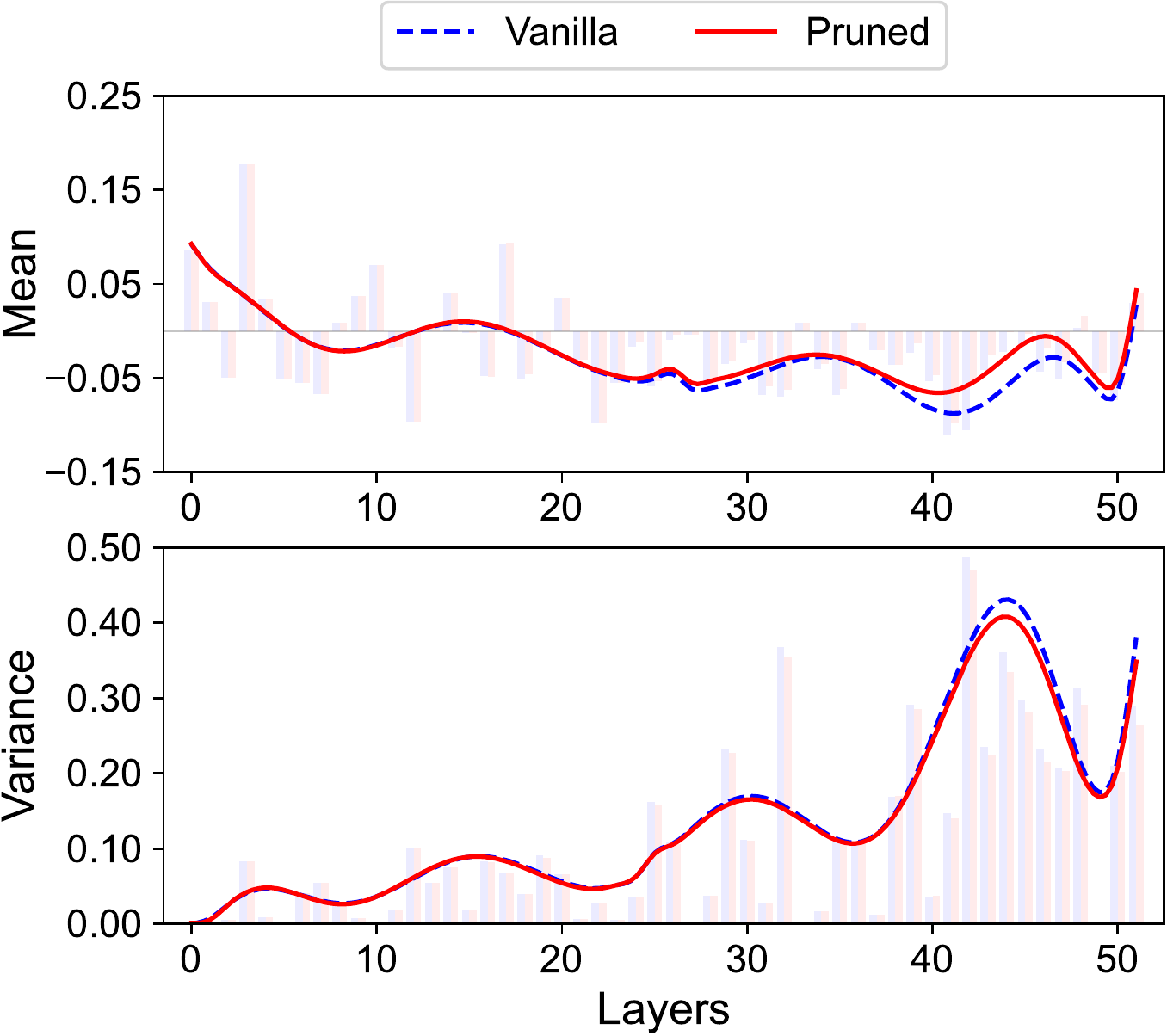}
\label{fig:variance}
\end{figure}

To answer these questions, we turn to analyze the pretrained DY-ResNet-50~\cite{he2015deep} model from the view of network pruning. Specifically, we prune out 50\% parameters of it with the lowest magnitude on CIFAR-100 dataset\footnote{https://github.com/weiaicunzai/pytorch-cifar100}~\cite{krizhevsky2009learning}. We measure the mean and variance values of the parameters in aggregated kernels as the proxy of the dynamic property: given different input samples, each dynamic convolution layer aggregates the computational kernel dynamically.By iterating over the entire validation dataset, we compute the layer-wise mean and variance of parameters in the aggregated kernel, which is shown in Figure~\ref{fig:variance}. Clearly, The change curves of the vanilla and the pruned networks almost coincide, with only some small divergences in the upper layers. Therefore, network pruning has little impact on the numeric features of the dynamic property. 

We also conduct a preliminary experiment to investigate the performance gap caused by network pruning on CIFAR-100 using ResNet~\cite{he2015deep} as backbones. We can see from results in Table \ref{tab:pre_exp} that pruned networks still maintain almost equally competitive performance: DY-ResNet families (with dynamic convolution layers) only encounter less than 1\% of performance drop and still outperform static networks by more than 2\% in accuracy. The above observations motivate us to explore effective and efficient sparse dynamic convolution structures. 

\begin{table}[t]
  \centering
  \caption{Preliminary results for network pruning in dynamic convolution (DY-Conv). After pruning $50\%$ parameters ``w/ Pruned'', dynamic convolutions still maintain comparable performance and the advantage over static convolutions ``Static''. \ding{86} indicates the dynamic models with the best performance, the fewest parameters, and the fewest FLOPs (``Static'' models are excluded). }
  \vspace{-10pt}
  \begin{center}
    \begin{adjustbox}{max width=\linewidth}
  \begin{tabular}{clrrc}
  \toprule
  Depth & Method & Param. & FLOPs & \makecell[l]{Top-1 (\%)} \\
    \midrule
   \multirow{3}{*}{ResNet-10} 
  & Static                  & 0.3M & 29.9M & 66.0 \\
  & DY-Conv                 & 1.2M & 34.8M & 68.9 \\
  & w/ Pruned                & \ding{86}0.6M & \ding{86}27.1M & 68.1\small{\textcolor{gray}{(-0.8)}} \\
  \midrule
  \multirow{3}{*}{ResNet-18} 
  & Static                  & 0.7M & 35.6M & 67.6   \\
  & DY-Conv                 & 2.8M & 43.4M & 72.4   \\
  & w/ Pruned                & \ding{86}1.4M & \ding{86}31.9M & 71.9\small{\textcolor{gray}{(-0.5)}}    \\
  \midrule
  \multirow{3}{*}{ResNet-50} 
  & Static                  & 1.5M & 122.3M & 72.2 \\
  & DY-Conv                 & 6.2M & 143.4M & 75.2 \\
  & w/ Pruned                & \ding{86}3.3M & \ding{86}108.5M  & 74.6\small{\textcolor{gray}{(-0.6)}} \\
  \toprule
  \end{tabular}
    \end{adjustbox}
  \end{center}
  \label{tab:pre_exp}
    \vspace{-10pt}
 \end{table}

\subsection{Sparse Dynamic Convolution}
In this section, we propose Sparse Dynamic Convolution, which utilizes parallel sparse kernels to aggregate dynamic kernels. We use binary masks $M$ to sparsify the kernels by pruning out unimportant parameters. Generally, a binary mask is a 0/1 matrix indexing the pruned weights in the parallel kernels. To make the binary mask trainable, we define a magnitude score ${S} = \| W\|$ and a threshold $\tau$. The mask is then rounded to 1 if the score is greater than the threshold, and vice versa, given by
\begin{equation}
 {M}_i = \begin{cases}1, & \text { if } {S}_i \ge \tau \\ 0, & \text { otherwise }\end{cases}.
 \label{eq_mask}
\end{equation}
The major challenge for training binary masks is that Eq.~(\ref{eq_mask}) is non-differentiable, impeding the calculation of gradients and blocking the updating process. To solve this problem, Piggyback~\cite{mallya2018piggyback} utilizes the Straight-Through Estimator (STE)~\cite{bengio2013estimating} (where the gradient is directly passed to its input $\frac{\partial \bm{M}}{\partial W} = \bm{1}$) to enable gradient estimation so that the gradient descent can update parameters.  

According to Zhou~\textit{et al.}~\cite{zhou2021learning}, the values of $\tilde{{M}}$ are not restricted in binary values 0/1 strictly, which may cause an unstable training process and accuracy drop. Inspired by Yang~\textit{et al.}~\cite{yang2020ksm}, we adopt the softmax function to approximate $\tilde{{M}}$ into binary values 0/1 for better gradient calculation: 
\begin{figure}[t]
  \centering
    \caption{Effect of the hyperparameter $T$ on the binary function Eq.~(\ref{eq:psi} and \ref{eq:mask}). It is easily observed that this hyperparameter contributes to the sharpness. By  decreasing $T$, we observe that  output values gradually follow an approximate 0/1 distribution. }
  \includegraphics[scale=0.4]{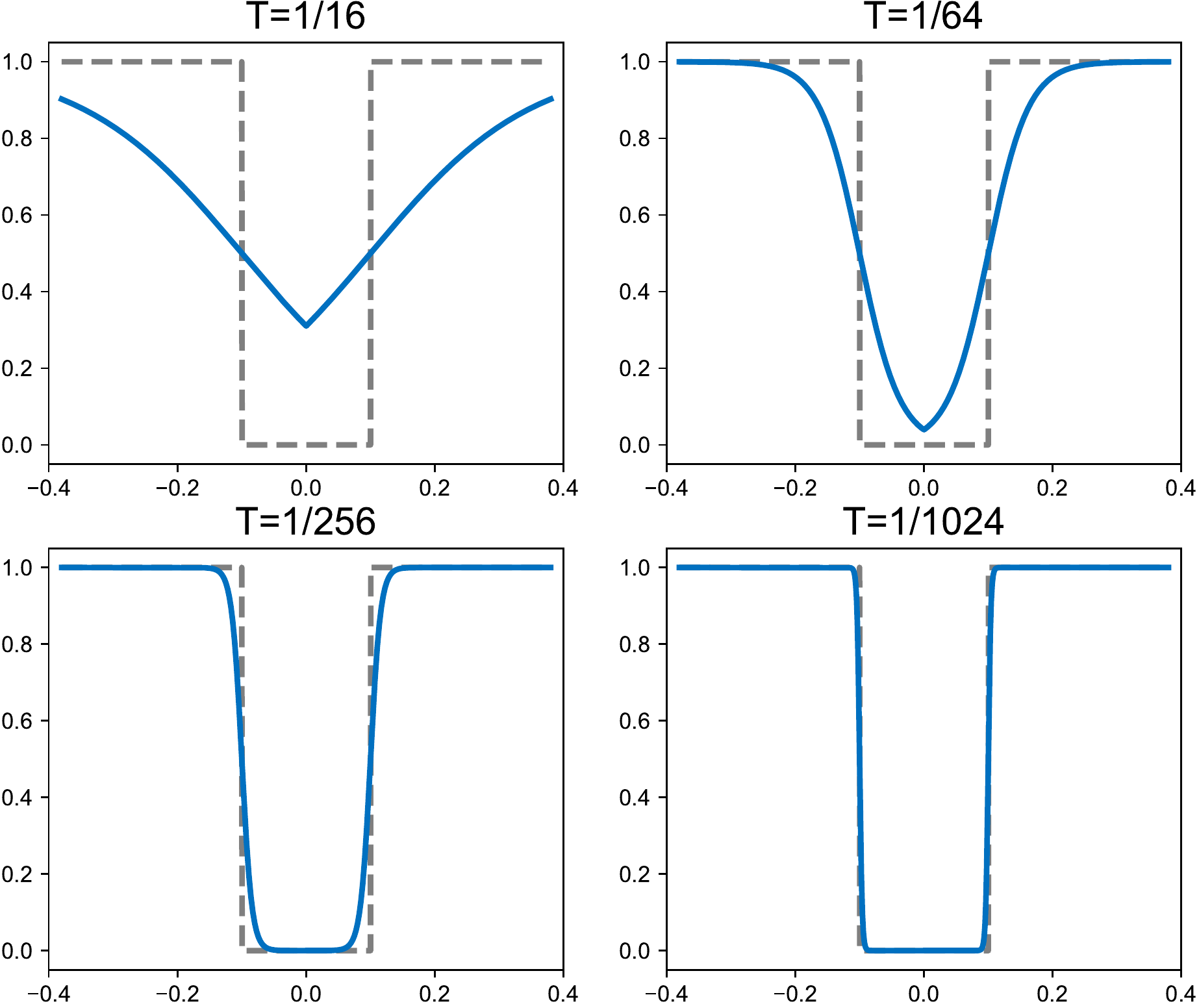}
\label{fig:function}
\end{figure}

\begin{equation}
    \psi_1 = \mathrm{\sigma}({S} - \tau),
    \label{eq:psi}
\end{equation}
\begin{equation}
     \tilde{{M}} = \frac {\mathrm{exp}(\psi_1/T)}{\mathrm{exp}(\psi_1/T) + \mathrm{exp}(\psi_0/T)},
     \label{eq:mask}
\end{equation}
where $\sigma(\cdot)$ is the sigmoid function and $\tau$ denotes the threshold. $\psi_0 = 1 - \psi_1$ is the complement of $\psi_1$, $\tilde M$ is the generated mask following an approximate 0/1 distribution. $T$ is the hyperparameter controlling the sharpness of the function. For example, $T=\frac{1}{1024}$ encourages the output to be either 0 or 1, which is shown in Figure \ref{fig:function}. Then we transform $\tilde{{M}}$ into binary values ${ M}$ using STE to generate and update the binary masks: 

 \begin{equation}
 {M} = \mathrm{round}({\tilde M}), \quad \nabla {{M}} = \nabla {\tilde{M}}.  
 \end{equation}
 
By utilizing binary masks to $k$ kernels, we transform the dynamic convolution into sparse dynamic convolution. In this layer, we first sparsify the $k$ parallel static convolution kernels ${W}_i$ $(i = 1,2, \dots, k)$ and then combine them dynamically, given by 

\begin{equation}
    \tilde{W}_i = {{M}}_i \odot W_i, \quad  \hat{W} = \sum_i  \pi_i * \tilde{W}_i.
\end{equation}
Here, $\tilde{W}_i$ and $\hat{W}$ denote the sparsified parameters of the static kernel and the aggregated sparse kernel, respectively. 

\subsection{Loss Function}
The binary mask ${ M}$ is determined by the magnitude of parameters $W$ and the threshold $\tau$. In general situations, $\tau$ is a hyperparameter as the threshold that controls the global sparsity. A naive way to set the threshold is to maintain the uniform sparsity of all layers. However, many experiments have indicated that setting multiple thresholds to control the layer-wise non-uniform sparsity performs much better \cite{kusupati2020soft, louizos2018learning, azarian2021learned}. Existing methods to acquire layer-wise sparsity are often dependent on hyperparameters and require iterative trials \cite{frankle2019lottery}. 

To address this problem, we propose a learning-based strategy to obtain layer-wise thresholds. Specifically, we first transform $\tau$ into learnable parameters and utilize it to generate differential masks: 
\begin{equation}
    \frac{\partial {  M}}{\partial \tau}  = 
    \frac{\partial {  M}}{\partial {\tilde M}}
    \frac{\partial {\tilde M}}{\partial {\psi}_1}
    \frac{\partial {\psi}_1}{\partial \tau}. 
\end{equation}
The gradient backpropagated to ${ M}$ indicates two directions: contributing to the performance improvement and constraining the overall sparsity. To constrain the sparsity, $L_{0}$-norm regularization has been widely researched in model sparsity~\cite{louizos2018learning, azarian2021learned}, for it directly regulates the overall parameter budget. Therefore, given the overall sparse level $s$, the non-zero ratio of overall parameters is $\bar s = 1 - s$. We resort to $L_{0}$-norm penalty $\mathcal{L}_s(\tau, s)$ to constrain the layer-wise non-uniform sparsity as followed: 
\begin{equation}
    \mathcal{L}_s(\tau, \bar s) = \mathrm{ReLU}(\sum_l N_l \cdot (|| M^{(l)}||_0 -  \bar s)), 
\end{equation}
where $\mathcal{L}_s$ is the regulation loss that controls the global sparsity, $N_l$ is the number of parameters in the $i$-th layer, ${{|| {{M}}}^{(l)}||_0}$ is the $L_{0}$ norm of the mask $ M$ in the $l$-th layer. Note that we use the ReLU function to restrict the global sparsity under the setting value, this loss term only works when the network is denser than expected. Formally, we define our loss function $\mathcal{L}$ as followed:
\begin{equation}
    \mathcal{L} = \mathcal{L}_c\big(y, f(\bm{x}, W, \tau)\big) + \lambda_s \mathcal{L}_s(\tau, \bar s) + \lambda_r ||W||_2,
\end{equation}
where we represent our networks as $f$ and the ground truth label as $y$. $\mathcal{L}_c$ is the standard loss function, e.g., cross-entropy in image classification. $||W||_2$ is the $L_2$ weight regularization loss and $\lambda_r$ is the weight decay rate. $\lambda_s$ is a hyperparameter that determines the pruning speed. 

\subsection{Optimization Policy}
We train the sparse dynamic convolution following an iterative pruning process \cite{frankle2019lottery, evci2021rigging}. Notably,  considering the time-consuming training process of dynamic networks, we restrain the total steps consistent with vanilla dynamic convolution and equally divide the total steps into $n + 1$ phases. Given the sparse level $s$ and pruning iterations $n$, in the first $n$ phases, we prune $s^{\frac{1}{n}}$ percent of the parameters at the end of each phase and retrain the network in the next phase. The whole training policy is shown in Algorithm~\ref{alg:training}. 

\begin{algorithm}[t]
  \caption{Sparse Dynamic Convolution} 
  \label{alg:training}
  \begin{algorithmic}[1]
    \REQUIRE
      Sparsity $s$, Total Steps $T$, Pruning Iterations $n$, \\ \hspace{-20pt} Dynamic Convolution Network $f$.
    \ENSURE
      Sparse Dynamic Network $f_s$.
    \STATE Initialize $\bar s = 1 - s$, $\bar{s}_0=0, \Delta t=\frac{T}{n + 1}$.
    \FOR{$t=1$ \rm{to} $T$}
      \STATE Compute loss $\mathcal{L} = \mathcal{L}_c + \lambda_s\mathcal{L}_s(\tau, \bar{s}_t) + \lambda_r ||W||_2$.
      \STATE Update parameters $W_{t+1} \leftarrow W_{t} - \eta_W \frac{\partial \mathcal{L}}{\partial W}$.
      \STATE Update thresholds $\tau_{t+1} \leftarrow \tau_t - \eta_{\tau} \frac{\partial \mathcal{L}}{\partial \tau}$.
      \IF{$t \operatorname{mod} \Delta t$ = $0$}
        \STATE Update variable $\begin{aligned} \bar{s}_{t + 1} = \bar{s}^{\frac{t}{\Delta t * n}} \end{aligned}$.
      \ELSE
        \STATE Pass variable $\begin{aligned} \bar{s}_{t+1} = \bar{s}_{t} \end{aligned}$.
      \ENDIF
    \ENDFOR{}
  \end{algorithmic}
\end{algorithm}

\section{Experiment}

In this section, we provide comprehensive experiments on both large-scale image recognition datasets and downstream tasks with different CNN architectures to validate the effectiveness of SD-Conv. Specifically, we compare the performance of sparse dynamic convolution with other convolution architectures, and further analyze the design of sparse dynamic convolution from the perspective of sparsity and reduced FLOPs.

\begin{table*}[h]
\caption{Comparison for MobileNetV2 and ResNet between Sparse Dynamic Convolution and baselines, including static convolution, Condconv\cite{yang2020condconv} and DY-Conv \cite{chen2020dynamic}. \ding{86} indicates the dynamic model with the fewest parameters or the fewest FLOPs (static models are not included). The best performance is \textbf{bold}. }
\vspace{-5pt}
\hspace{-14pt}
\begin{minipage}{\linewidth}
\centering
  \begin{minipage}[t]{0.46\textwidth}
    \begin{tabular}{clrrc}
      \toprule
      Width & Method & Param. & FLOPs & \makecell[l]{Top-1 (\%)} \\
      \midrule
      \multirow{4}{*}{$\times 1.0$} 
      & Static & 3.5M & 300.0M & 72.0\\
      & CondConv & 27.5M & 329.0M & 74.6 \\
      & DY-Conv & 11.8M & 312.9M & 75.2 \\
      & SD-Conv & \ding{86}7.7M & \ding{86}261.9M & \textbf{75.3}    \\
      \midrule
      \multirow{4}{*}{$\times 0.75$} 
      & Static & 2.6M & 209.1M & 69.3 \\
      & CondConv & 17.5M & 233.9M & 71.8\\
      & DY-Conv & 7.6M & 220.1M & 72.8 \\
      & SD-Conv & \ding{86}5.0M & \ding{86}171.8M & \textbf{73.2} \\
      \midrule
      \multirow{4}{*}{$\times 0.5$} 
      & Static & 2.0M & 97.0M & 65.4\\
      & CondConv & 15.5M & 113.0M & 68.4 \\
      & DY-Conv & 4.4M & 101.4M & 69.9 \\
      & SD-Conv & \ding{86}3.1M & \ding{86}81.5M & \textbf{70.3} \\
      \toprule
    \end{tabular}
    \subcaption{\normalsize \bf MobileNetV2}
    \label{tab:mbv2}
  \end{minipage}\hspace{15pt}\begin{minipage}[t]{0.46\textwidth}
    \begin{tabular}{clrrc}
      \toprule
      Depth & Method & Param. & FLOPs & \makecell[l]{Top-1 (\%)} \\
      \midrule
      \multirow{4}{*}{ResNet-10} 
      & Static & 5.2M & 0.89G & 63.4 \\
      & CondConv & 36.7M & 0.92G & 66.8 \\
      & DY-Conv & 18.6M & 0.91G & 67.5 \\
      & SD-Conv & \ding{86}10.4M & \ding{86}0.73G & \textbf{67.9} \\
      \midrule
      \multirow{4}{*}{ResNet-18} & Static & 11.1M & 1.81G & 70.4\\
      & CondConv & 81.4M & 1.89G & 72.0\\
      & DY-Conv & 42.7M & 1.85G & 72.7\\
      & SD-Conv & \ding{86}23.2M & \ding{86}1.51G & \textbf{73.3}    \\
      \midrule
      \multirow{4}{*}{ResNet-50} 
      & Static & 23.5M & 3.8G & 76.2 \\
      & CondConv & 129.9M &4.0G & 76.8\\
      & DY-Conv & 100.9M & 4.0G& 77.3 \\
      & SD-Conv & \ding{86}54.0M & \ding{86}3.4G  & \textbf{77.4} \\
      \toprule
    \end{tabular}
    \subcaption{\normalsize \bf ResNet}
    \label{tab:resnet}
  \end{minipage}
\end{minipage}
\vspace{-10pt}
\end{table*}

\subsection{Image Classification on ImageNet}

Our main experiments are implemented on the ImageNet dataset \cite{5206848}, which is one of the most challenging image classification datasets with 1,000 classes, including 1,281,167 images for training and 50,000 images for validation. 

\textbf{CNN Backbones.} We use ResNet \cite{he2015deep} and MobileNetV2 \cite{sandler2019mobilenetv2}  families for experiments, covering both light-weight CNN architectures and larger ones. Specifically, we choose ResNet-10, ResNet-18, ResNet-50 and MobileNetV2 ($1.0\times, 0.75\times, 0.5\times$) as the backbones. 

\textbf{Experimental Setup.} We validate the effectiveness of our method by replacing dynamic convolution for all convolution layers except the first layer. Each layer has $k = 4$ experts with the reduce ratio as $16$ for the attention block in dynamic convolution \cite{chen2020dynamic}. We use an SGD optimizer \cite{ruder2016overview} with 0.9 momentum, following cosine learning rate scheduling and warmup strategy. The learning rate rises to the max learning rate linearly in the first 10 epochs and is scheduled to arrive at zero within a single cosine cycle. When generating binary masks, we set constant $T=\frac{1}{1024}$ to ensure $\tilde M$ follows approximately 0/1 binary values. The scale factor $\lambda_s$ of sparse penalty $\mathcal{L}_s(\tau, s)$ is fixed as 0.01. We follow Zhou~\textit{et al.}~\cite{chen2020dynamic}'s temperature annealing strategy to avoid the unstable output values of the softmax function in the first epoch. We train the ResNet models for 100 epochs, and the max learning rate is 0.1. For the MobilenetV2 models, we train them for 300 epochs, and the max learning rate is 0.05. The weight decay is 4e-5 for all models. 

\textbf{Main Results.} Table \ref{tab:mbv2} and \ref{tab:resnet} show the comparison between SD-Conv and other convolution architectures in two CNN architectures (ResNet and MobilenetV2). Our baselines include the static convolution, CondConv \cite{yang2020condconv} and DY-Conv \cite{chen2020dynamic}. We set $s=50\%$ to make the overall sparsity over 50\%. As shown, sparse dynamic convolution achieves significant performance improvement with a much smaller model size compared to vanilla dynamic convolution. For ResNet-18, sparse dynamic convolution has only 54.3\% of the parameters of vanilla dynamic convolution. For MobilenetV2-1.0, our method only requires 53.5\% of the parameters of dynamic convolution to achieve the same level of accuracy. The most prominent advantage of sparse dynamic convolution is its low computational cost. Owing to the sparse computational kernel $\hat W$, our method requires much fewer FLOPs in the convolution operation that acts as the dominant part of the overall FLOPs. The computational cost of our method is even less than that of static convolution, while all the other dynamic networks introduce extra computational costs. For example, sparse dynamic convolution only has 87.3\% of FLOPs of static convolution in MobilenetV2-1.0. 

\textbf{Robustness.} Traditional network structures are robust to the images perturbed with small Gaussian noise, while networks pruned with random masks can even have higher robustness than normal ones \cite{luo2020random}. To check whether SD-Conv also enjoys such property or even has better robustness in this scenario, we also consider the model's endurance of noise attack. We conduct an robustness evaluation on ImageNet for ResNet-50 \cite{he2015deep}.  Inspired by Luo~\textit{et al.}~'s work \cite{luo2020random}, we feed input images with Gaussian noises $z \sim N\left(0, \sigma^{2}\right)$ to networks. Table \ref{robustness} shows the robustness evaluation on random noise attack, we set the sparse ratio $s$ as 20\% and 80\% separately for our model. Disturbed by the same intensity of noise, we can see that our networks have the highest accuracy and gain up to 0.5\% improvement compared to dynamic architectures.  

\begin{table}[t]
\caption{The robustness evaluation based on random noise attack. Setting different standard variance $\sigma$, we evaluate the performance of different models. The best performance is \textbf{bold}. }
        \vspace{-5pt}
  \begin{adjustbox}{max width=\linewidth}
  \centering
  \begin{tabular}{clcccc}
  \toprule
  Model & Option & 0.05  & 0.10 & 0.15 & 0.20  \\ \toprule
   \multirow{3}{*}{ResNet-50}  & Static & 68.2 & 65.4 & 58.4 & 54.2 \\ 
                               & DY-Conv & 68.7 & 66.1 & 59.2 & 55.4 \\
                               & SD-Conv & \bf 69.1 & \bf 66.5 & \bf 59.5 & \bf 55.9 \\
                        
  \midrule               
   \multirow{3}{*}{ResNet-18}  & Static & 60.7 &  53.9 & 49.8 & 45.1 \\
                               & DY-Conv & 61.1 & 54.8 & 50.4 & 46.3 \\
                               & SD-Conv & \bf 61.3 & \bf 55.2 & \bf 50.5 & \bf 46.7 \\ \toprule
  \end{tabular}
  \end{adjustbox}
\label{robustness}
\end{table} 

\subsection{Transferring to Downstream Tasks}
\begin{figure*}[t]
  \centering
    \caption{Transferability comparison between Static, Dynamic, and our proposed SD-Conv from pretrained ResNet to different downstream tasks. We report two finetuning approaches: linear finetuning and full finetuning. }
    \vspace{-5pt}
  \includegraphics[scale=0.5]{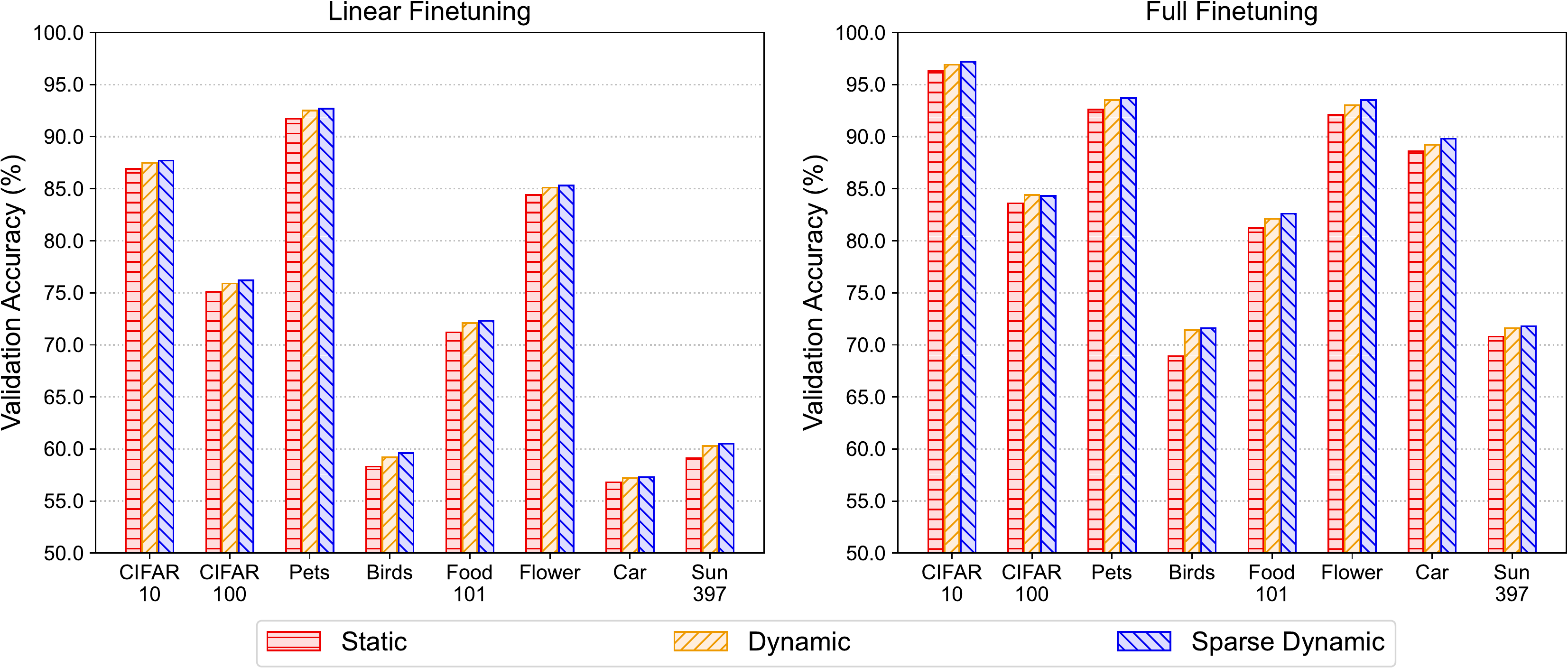}
\vspace{-15pt}
 \label{fig:down_stream}
\end{figure*} 

Network architectures measured against ImageNet \cite{5206848} have fueled much progress in computer vision researches across a broad array of problems, including transferring to new datasets \cite{dong2018domain, sharif2014cnn}, object detection \cite{huang2017speed}, image segmentation \cite{he2017mask} and perceptual metrics of images \cite{johnson2016perceptual}. Many previous works have proved that better network architectures learn better features to be transferred across vision-based tasks \cite{howard2017mobilenets, sharif2014cnn}. Therefore, we further evaluate the effectiveness of our network on downstream vision tasks, including CIFAR-10 \cite{krizhevsky2009learning}, CIFAR-100 \cite{krizhevsky2009learning}, Oxford-IIIT Pets \cite{parkhi2012cats}, Birdsnap \cite{berg2014birdsnap}, Food-101 \cite{bossard2014food}, Oxford 102 Flowers \cite{nilsback2008automated}, Stanford Cars \cite{krause2013collecting}, SUN397 \cite{xiao2010sun}. These tasks span several domains, difficulties, and data sizes. 

We transfer all parameters of the upstream model except the last (fully connected) layer, which is adjusted to the number of classes in the downstream task, using Kaiming uniform initialization \cite{he2015delving}. We finetune the pretrained ImageNet model following two strategies, linear finetuning and full finetuning. For linear finetuning, we only train the linear classifier ``on the top" of a fixed representation on downstream tasks, while we re-initialize the final layer and train the whole model for full finetuning. For both strategies, we take top-1 classification accuracy as the metric to compare different structures, which is shown in Figure~\ref{fig:down_stream}. The results clearly show that the sparse dynamic convolution achieves consistent improvement compared to dynamic convolution on downstream tasks, suggesting that pruning redundant information in the weights is beneficial to dynamic convolution architectures in transfer settings. 

\begin{figure} [ht]
  \centering
  \caption{The ablation study on sparsity for ResNet. Dotted line represents the performance of static convolution.}
  \begin{subfigure}{\linewidth}
  \includegraphics[scale=0.43]{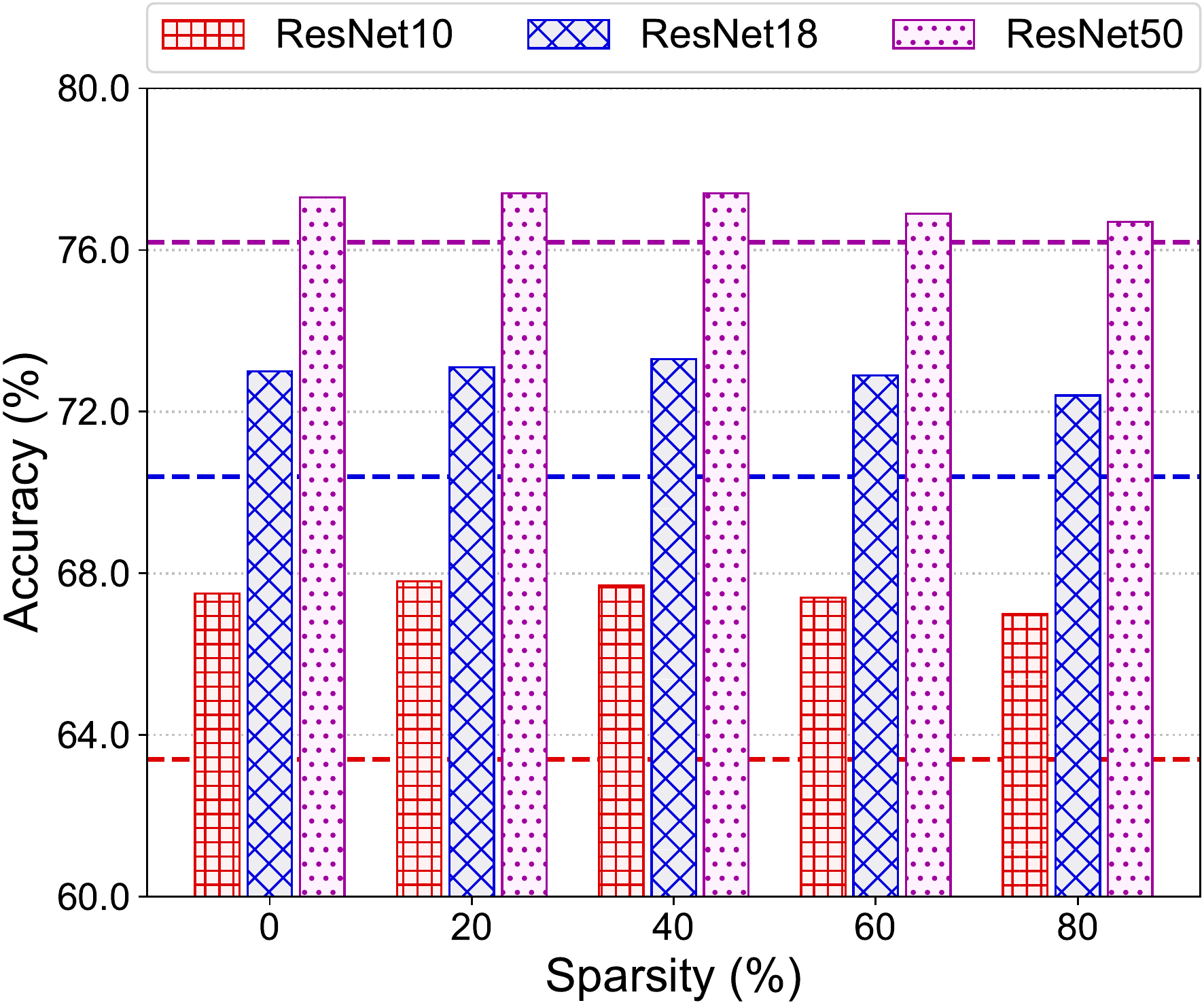}
 \end{subfigure}
 \vspace{-15pt}
 \label{fig:sparsity}
\end{figure}

\subsection{Further Analysis}
\paragraph{Further Increased Sparsity Can Still Maintain Superior Performance.}
To further explore the impact of sparsity, we conduct an ablation study by investigating a series of sparse ratios (from 20\% to 80\%). Figure \ref{fig:sparsity} shows the result of the ablation study on ImageNet classification \cite{5206848} experiments for ResNet \cite{he2015deep} in different depth, where we directly report the classification accuracy. For ResNet models with different depths, we can observe a consistent phenomenon that SD-Conv performs stably under different degrees of network pruning. At low sparse ratios (e.g. $s\le40\%$),  pruning out some unimportant parameters can lead to higher performance. When further increasing the sparse ratios, e.g. 60\% and 80\%, sparse dynamic convolution networks still maintain a significant performance advantage over static convolution networks. Considering the competitive performance, network pruning is an efficient way to simplify and promote dynamic convolution. 

\begin{figure} [t!]
  \centering
  \caption{The curves of kernel and layer sparsity for ResNet-18.}
  \includegraphics[scale=0.5]{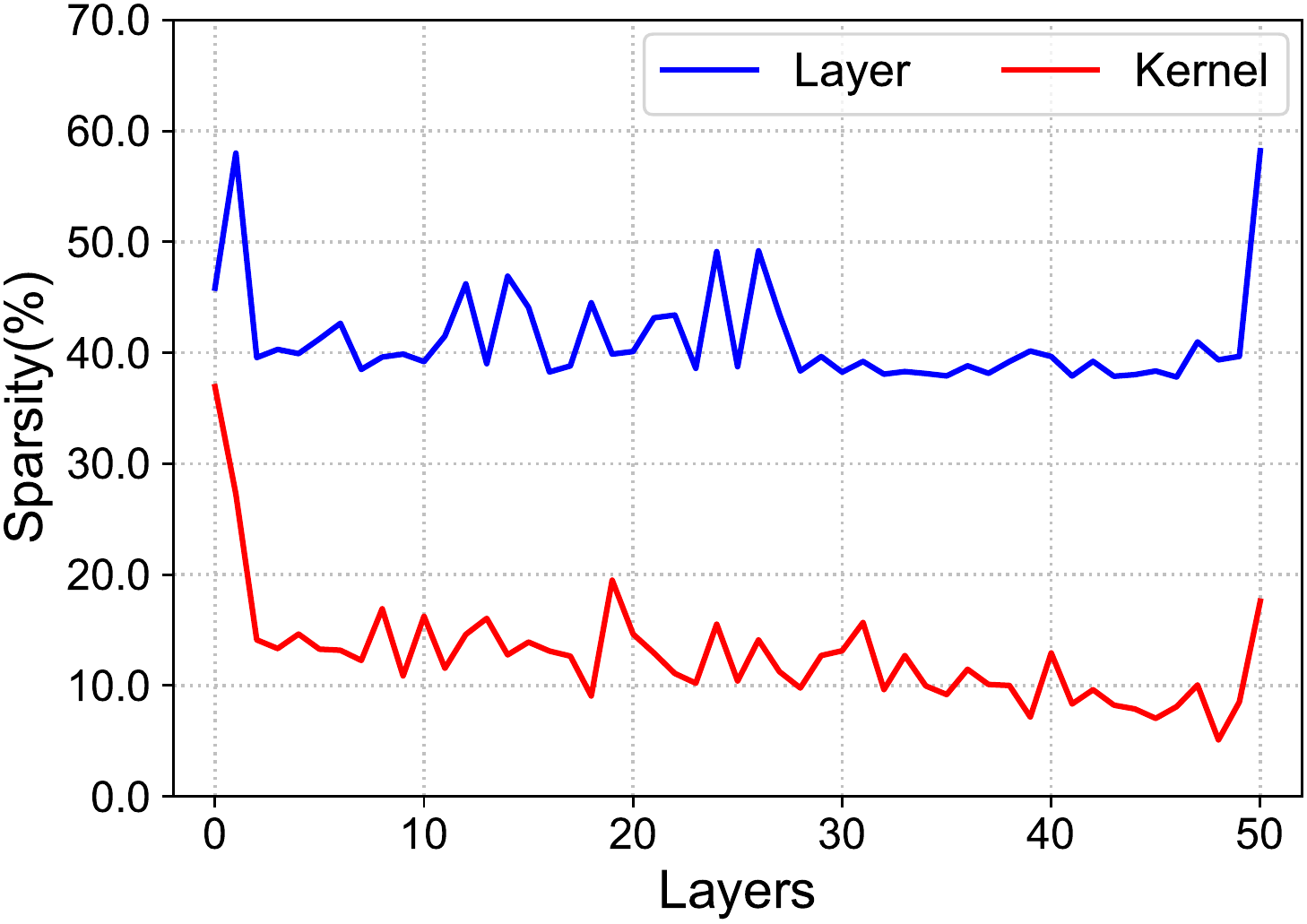}
 \label{fig:distribution}
\end{figure}

\paragraph{Reduced FLOPs Come from the Sparse Aggregated Kernel.}
\label{sec:reduced}
As shown in Table~\ref{tab:mbv2} and Table~\ref{tab:resnet}, the FLOPs of sparse dynamic convolution networks are even lower than static convolution. According to our observation, the reduced FLOPs come from the sparsity in propagated kernels $\hat W$. Even so, we have to mention that the sparsity of $\hat W$ is not definitely dependent on the sparsity of parameter $\tilde W_j$ $(j = 1,2, \dots, k)$ but lies in the overlap between them: the $i$-th element of $\hat W$ is zero only when all static kernels $\tilde W_j$ $(j = 1,2, \dots, k)$ have zero elements in the $i$-th position. Therefore, other than layer sparsity towards parameters (the proportion of zero-elements in $\tilde W_j$), we also resort to the kernel sparsity towards propagated kernels (the proportion of zero-elements in $\hat{w}$). To investigate the distribution of pruned parameters in propagated kernels, we visualize the pruned ratio of propagated kernels $\hat{W}$ and $k$ kernels $W_k$ in Figure \ref{fig:distribution}. We show that the kernel sparsity follows a similar trend to layer sparsity but maintains relatively smaller values. Even so, each propagated kernel still maintains a certain degree of sparsity, and the pruned weights contribute to the reduced FLOPs compared to dense convolution kernels. 

\section{Discussion of Masking Strategy}

\begin{table}[htbp]
  \centering
  \caption{Comparison between two different masking strategies. We use ``Diff'' to denote the different masking strategy and ``Same'' to denote the same masking strategy. \ding{86} indicates the dynamic model with the fewest parameters or the fewest FLOPs. 
  }
  \begin{adjustbox}{max width=\linewidth}
  \begin{tabular}[c]{llllll}
  \toprule
  Network & Method & Param & Flops & Acc(\%) \\
  \midrule
  \multirow{4}{*}{ResNet-50} & Static & 2.35M & 3.8G & 76.2 \\
  & DY-Conv & 100.9M & 4.0G & 77.3 \\
  & Diff & \ding{86}63.3M & 3.5G & \bf 77.4 \\
  & Same & \ding{86}63.3M & \ding{86}2.5G & 76.6 \\
  \midrule
  \multirow{4}{*}{MobilenetV2-1.0} & Static & 3.5M & 300.0M & 72.0 \\
  & DY-Conv & 11.1M & 312.9M & 75.2 \\
  & Diff & \ding{86}5.3M & 271.9M & \bf 75.3 \\
  & Same & \ding{86}5.3M & \ding{86}192.3M & 74.6 \\
  \toprule
  \end{tabular}
  \end{adjustbox}
  \label{tab:diff}
  \end{table}
  
\begin{figure}[t]
\centering
\caption{Comparison between two masking strategies. The left one is the default setting in SD-Conv, which takes $k$ different masks for each counterpart kernel. On the right, each kernel shares the same mask.}
\vspace{-10pt}
\includegraphics[width=\linewidth]{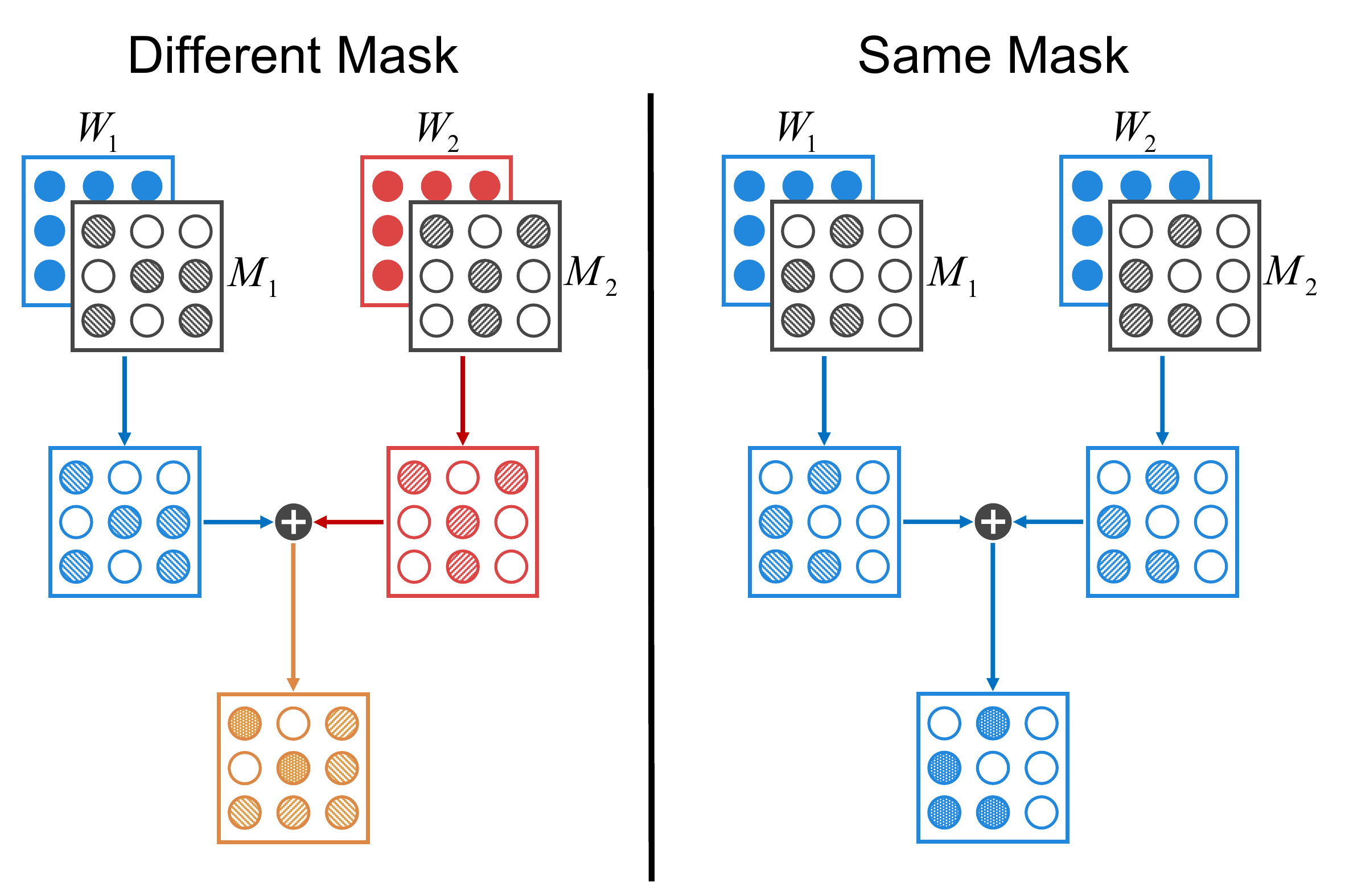}
\vspace{-20pt}
\label{fig:mask}
\end{figure}

As aforementioned in Section \ref{sec:reduced}, the kernel sparsity lies in the overlap region of $k$ masks ${M}_i$ ($i = 1,2, \dots, k$) and is usually lower than the parameter sparse ratio. Only when ${M}_1 = {M}_2 = \dots = {M}_k$, the kernel sparsity can be the highest and the FLOPs can be minimized, . Therefore, we evaluate a strategy that directly applies the same mask to the static kernels and then compare it with our proposed method, as shown in Figure \ref{fig:mask}. We can see from the numeric results from Table \ref{tab:diff} that utilizing the same mask to $k$ kernels can cause a performance drop though it significantly reduces the FLOPs. In contrast, our learning-oriented thresholds lead to different masks among static kernels and obtain significantly better results. We believe that the sparser aggregated kernels cause the performance drop and there exists a trade-off between optimal FLOPs and performance in sparse dynamic convolution. 

\section{Conclusion}
In this work, we systematically re-examine the parameter efficiency property of dynamic convolution networks through the lens of network pruning. Based on our findings, we propose a plug-in strategy, i.e. Sparse Dynamic Convolution, for existing dynamic convolution methods. Our method improves the performance of dynamic convolution both in upstream ImageNet classification and a variety of downstream tasks, with fewer parameters and FLOPs. Our study empirically indicates the effectiveness of sparsity in dynamic convolution and informs the potential to further promote sparse dynamic convolution in view of the trade-off between performance and FLOPs.

In future work, we would like to investigate the parameter efficiencies of other neural network models, especially for scenarios where high efficiency is required, e.g., Adapter~\cite{he-etal-2022-sparseadapter}, Prompt~\cite{Zhong2022PANDAPT} and the distilled student~\cite{Rao2022ParameterEfficientAS}.

\section*{Acknowledgements}
We are grateful to the anonymous WACV reviewers and the area chair for their insightful comments and suggestions.

{\small
\bibliographystyle{ieee_fullname}
\bibliography{egbib}
}

\end{document}